\documentclass[conference]{IEEEtran}
\IEEEoverridecommandlockouts
\usepackage{cite}
\usepackage{amsmath,amssymb,amsfonts}
\usepackage{algorithmic}
\usepackage{graphicx}
\usepackage{textcomp}
\usepackage{xcolor}
\usepackage{booktabs}
\usepackage{multirow}
\def\BibTeX{{\rm B\kern-.05em{\sc i\kern-.025em b}\kern-.08em
    T\kern-.1667em\lower.7ex\hbox{E}\kern-.125emX}}
\begin{document}

\title{Do Self-Supervised Speech Models Exhibit\\
the Critical Period Effects in Language Acquisition?
\thanks{This work was partially supported by the ``R\&D Hub Aimed at Ensuring Transparency and Reliability of Generative AI Models'' project of the Ministry of Education, Culture, Sports, Science and Technology.}
}

\author{\IEEEauthorblockN{Yurie Koga$^1$ \qquad Shunsuke Kando$^1$ \qquad Yusuke Miyao$^{1,2}$} \IEEEauthorblockA{\textit{$^1$Department of Computer Science, The University of Tokyo, Japan \qquad $^2$NII LLMC, Japan} \\
\{ykrasp7isweet, skando, yusuke\}@is.s.u-tokyo.ac.jp}
}

\maketitle

\IEEEpubid{\begin{minipage}{\textwidth}\vspace{80pt}
\centering \scriptsize
© 2025 IEEE.  Personal use of this material is permitted. Permission from IEEE must be obtained for all other uses, in any current or future media, including reprinting/republishing this material for advertising or promotional purposes, creating new collective works, for resale or redistribution to servers or lists, or reuse of any copyrighted component of this work in other works.
\end{minipage}}

\begin{abstract}
This paper investigates whether the Critical Period (CP) effects in human language acquisition are observed in self-supervised speech models (S3Ms).
CP effects refer to greater difficulty in acquiring a second language (L2) with delayed L2 exposure onset, and greater retention of their first language (L1) with delayed L1 exposure offset.
While previous work has studied these effects using textual language models, their presence in speech models remains underexplored despite the central role of spoken language in human language acquisition.
We train S3Ms with varying L2 training onsets and L1 training offsets on child-directed speech and evaluate their phone discrimination performance.
We find that S3Ms do not exhibit clear evidence of either CP effects in terms of phonological acquisition.
Notably, models with delayed L2 exposure onset tend to perform better on L2 and delayed L1 exposure offset leads to L1 forgetting.
\end{abstract}


\section{Introduction}
These days, self-supervised speech models (S3Ms) achieve performance comparable to humans on various language-related tasks~\cite{yang-etal-2024-large}.
This has led to the increase of studies aiming to gain insights into human language acquisition mechanism by comparing the learning process of these models with that of humans~\cite{matusevych-etal-2021-phonetic,zaitova-etal-2022-mapping,lavechin-etal-2024-modeling,millet-dunbar-2022-self}.
By examining whether computational models exhibit human-like behaviors, researchers aim to determine whether such human behaviors are the results of statistical learning itself, or whether additional innate mechanisms account for such human behaviors.

One of the famous phenomena in human language acquisition is the Critical Period (CP) phenomenon.
This refers to the idea that there is a specific period in human development, known as the CP, when language can be acquired effectively, and that this ability gradually decreases after this period~\cite{lenneberg-1967-biological}.
This effect is observed in the acquisition of various language abilities, such as phonological, syntactic, and semantic acquisition~\cite{singleton-2005-critical}.
However, the underlying mechanism of this phenomenon remains unclear.
While some theories attribute it to an innate decline in neural plasticity governed by biological maturation~\cite{lenneberg-1967-biological,pinker-1994-language}, others argue that the process of language learning itself induces the reduction in plasticity that causes the CP phenomenon~\cite{munro-1986-state,elman-etal-1996-rethinking}.

\begin{figure}[tb]
    \centerline{\includegraphics[width=8.7cm]{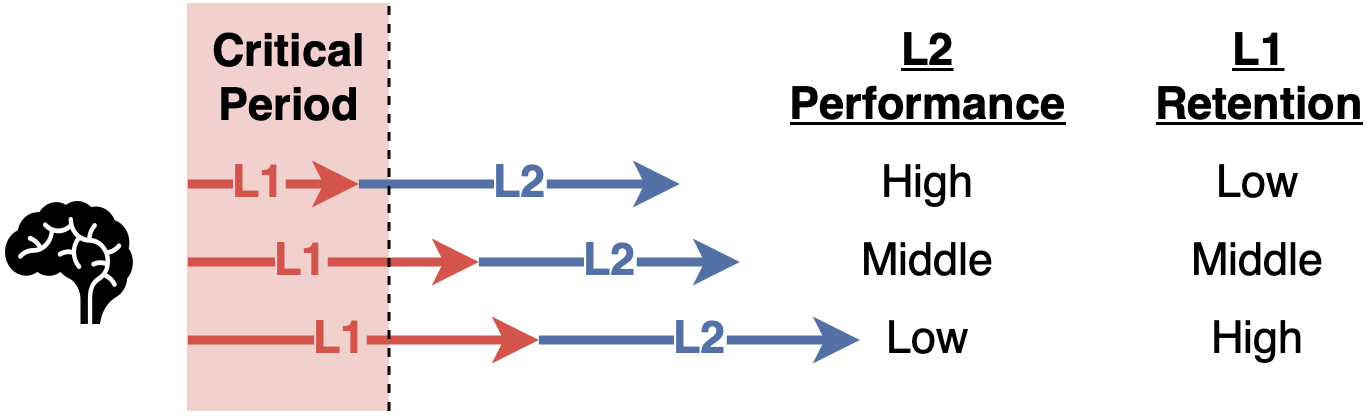}}
    \caption{Overview of the Critical Period (CP) effects in human language acquisition focused in this study. L2 performance shows the CP effects for L2 acquisition, and L1 retention shows the effects for L1 attrition. We examine these effects on HuBERT.}
    \label{fig:CP}
\end{figure}

In this study, we focus on the CP effects for second language (L2) acquisition and first language (L1) attrition (Fig.~\ref{fig:CP}), in terms of phonological acquisition.
The former effect is characterized by the requirement of early L2 exposure before the end of CP for attaining native-like proficiency, and increasing difficulty in L2 acquisition as the onset of L2 exposure is delayed.
In contrast, the latter effect suggests that L1 is forgotten if L1 exposure ceases before the end of CP, while prolonged exposure reduces the likelihood of forgetting it.
These effects are known to be not reproduced in text language models when trained without any explicit modeling of the decrease of neural plasticity, which aligns with the innateness hypothesis~\cite{constantinescu-etal-2024-investigating}.
However, relatively little attention has been paid to speech-based approaches, although speech plays a central role in early language learning and more closely reflects the conditions under which children naturally acquire language.

To address this problem, we investigate whether the CP effects for L2 acquisition and L1 attrition are observed in self-supervised speech models (S3Ms).
Specifically, we train HuBERT~\cite{hsu-etal-2021-hubert} with various L2 training onsets and L1 training offsets, and evaluate their phone discrimination performance for each language.
We utilize child-directed datasets (CDS) for training to more accurately replicate the language acquisition process in humans.
Regarding the CP effects for L2 acquisition, we examine whether later L2 exposure onset results in lower L2 phone discrimination ability.
Similarly, regarding L1 attrition, we investigate whether earlier L1 exposure offset leads to a more severe decrease in L1 phone discrimination ability.

We find that CP effects are not observed in HuBERT for either L2 acquisition or L1 attrition.
Specifically, a delayed onset of L2 exposure led to improved L2 performance, and a delayed offset of L1 exposure resulted in L1 forgetting, while models with an earlier L1 exposure offset showed improved L1 performance with subsequent L2 training.
This result indicates that CP effects are not a necessary consequence of statistical learning itself, which is consistent with the innateness hypothesis.
\section{Background and Related Work}
\subsection{The Critical Period}
The CP hypothesis proposes that the efficiency of human language acquisition declines gradually after a specific period (CP), during which language can be acquired with relative ease~\cite{lenneberg-1967-biological}.
A wide range of empirical studies have reported CP effects in humans that support this hypothesis across various environmental conditions~\cite{newport-1990-maturational,pallier-2007-critical,asher-garcia-1969-optimal}, and the hypothesis is generally accepted as valid.
Although the exact timing of the CP offset and the nature of the decline in ultimate language proficiency remain debated, CP effects are said to apply across multiple domains of language ability, including phonology, syntax, and semantics~\cite{singleton-2005-critical}, regardless of the first and second language~\cite{hartshorne-etal-2018-critical}.
CP effects commonly consist of three effects: one related to L1 acquisition, one to L2 acquisition, and one to L1 attrition.
In this study, we focus on the latter two---L2 acquisition and L1 attrition---as the CP effects for L1 acquisition are difficult to examine with computational models.

\subsubsection{CP effects for L1 acquisition}
This phenomenon refers to the observation that individuals who are not exposed to their L1 within CP fail to attain full native-like L1 proficiency.
In particular, the ultimate level of L1 proficiency tends to decline as the age of initial L1 exposure increases.
A well-studied example of this phenomenon comes from research on deaf individuals who acquired American Sign Language (ASL).
Those who were exposed to ASL from birth or early childhood outperformed late learners in a morphology-related task~\cite{newport-1990-maturational}.

\subsubsection{CP effects for L2 acquisition}
This phenomenon refers to the increasing difficulty of acquiring an L2 as the onset of L2 exposure is delayed.
This pattern has been observed across different linguistic domains, including phonology and grammar.
For example, early L2 learners are more likely to achieve a native-like accent, whereas those who begin learning in adulthood typically can not~\cite{asher-garcia-1969-optimal, seliger-etal-1975-maturational, oyama-1976-sensitive}.
In addition, in terms of grammatical ability, native Korean and Chinese immigrants to the United States showed a clear age-related decline in performance on grammaticality judgment tasks~\cite{johnson-newport-1989-critical}.
A large-scale study with participants from all over the world also confirmed the existence of CP effects in L2 grammatical acquisition~\cite{hartshorne-etal-2018-critical}.

\subsubsection{CP effects for L1 attrition}
This phenomenon refers to the observation that individuals exposed to their L1 only before the end of CP (in early childhood) are more likely to forget L1 than those exposed to their L1 beyond CP.
For instance, Korean-born individuals adopted by French families before the age of 10 were found to completely forget Korean, as evidenced by their poor performance on word meaning recognition, sentence identification, and speech segment detection tasks~\cite{pallier-2007-critical}.

\subsection{Testing CP effects with computational models}
The cause of the CP phenomenon was originally attributed to innate neurobiological factors, particularly the completion of the brain lateralization, in which the language functions become dominant in the left hemisphere of the brain~\cite{lenneberg-1967-biological}.
While this nativist view posits that the decline in plasticity is genetically programmed, an alternative empirical hypothesis suggests that the act of language learning itself reduces this plasticity over time~\cite{munro-1986-state}.
One promising approach to distinguish between these explanations is to examine whether CP effects can emerge purely from the statistical learning of computational models.
Such models allow precise control over both the quality and timing of language exposure.
Recent work has explored this using text-based language models~\cite{constantinescu-etal-2024-investigating, mita-etal-2025-developmentally}, but these do not fully capture the conditions of early human language acquisition, as infants first encounter language through speech rather than written text.

\subsection{Comparison of L2 perception between speech models and humans}
Several studies have examined whether speech models exhibit human-like behavior in L2 perception, though their focus is different from CP effects.
Early work using RNN-based models explored whether models perform better on the L2 phone discrimination task when trained with a higher proportion of L2 data~\cite{matusevych-etal-2021-phonetic}, or whether they can recognize L2 word meanings without direct exposure to L2 when it closely resembles their L1~\cite{zaitova-etal-2022-mapping}.
More recently, self-supervised speech models (S3Ms) such as Contrastive Predictive Coding (CPC)~\cite{oord-etal-2019-representation}, HuBERT~\cite{hsu-etal-2021-hubert}, and wav2vec 2.0~\cite{baevski-etal-2020-wav2vec2.0} have attracted attention for modeling human perception due to their strong performance across a range of tasks.
For example, CPC has been tested for its ability to discriminate phonemes absent in the model's training language during training~\cite{lavechin-etal-2024-modeling}.
This study revealed that the decrease of this ability like humans emerges only when trained on clean speech, but not when trained on noisy child-directed speech (CDS).
In contrast, HuBERT and wav2vec 2.0 did not exhibit native language effects in phone discrimination tasks when trained on clean speech~\cite{millet-dunbar-2022-self}, though it has not been tested on CDS.
\section{Experiment}
In this study, we investigate whether S3Ms exhibit CP effects by manipulating the timing of exposure to L2 during training.
Specifically, we address the following two questions: 
(1) CP for L2 acquisition: Does delayed onset of L2 exposure, given equal L2 training duration, lead to lower L2 phone discrimination performance? 
(2) CP for L1 attrition: Does earlier offset of L1 exposure, given equal L2 training duration, result in a more severe decline in L1 phone discrimination performance?
To examine the potential effects of similarity between L1 and L2, we constructed two language pairs: (English, Japanese) and (English, French), with the latter being more phonologically similar.
For each pair, we conducted experiments in both directions, with each language serving as either L1 or L2.
All evaluations were conducted in English due to the availability of a large evaluation dataset.
We also experimented with two L2 training settings: training with an L2-only dataset (L2-only) and training with an L1-L2-mixed dataset (L1+L2), which aims to realize human-like learning.
The overall experimental settings and their correspondence to the two CP effects are shown in Table~\ref{tab:exp-setting}.
\begin{table}[tb]
    \centering
    \renewcommand{\arraystretch}{1.3}
    \caption{Experimental settings and their correspondences with CP effects. We test with four (L1, L2) pairs and two L2 training settings. The L2 training setting means whether to train the model in the order of L1$\rightarrow$L2 (L2-only) or L1$\rightarrow$L1+L2 (L1+L2). We use only English for evaluation.}
    \label{tab:exp-setting}
    \begin{tabular}{l|cl|cl}
        \multirow{2}{*}{\begin{tabular}[c]{@{}l@{}}\textbf{L2 Training}\\\textbf{Setting}\end{tabular}} & \multicolumn{4}{c}{\textbf{(L1, L2)}} \\ \cline{2-5}
        & \multicolumn{1}{l}{(JA, EN)} & (FR, EN) & \multicolumn{1}{l}{(EN, JA)} & (EN, FR) \\
        \hline
        L2-only & \multicolumn{2}{c|}{\multirow{2}{*}{\begin{tabular}[c]{@{}c@{}}CP for\\ L2 acquisition\end{tabular}}} & \multicolumn{2}{c}{CP for L1 attrition} \\ \cline{1-1} \cline{4-5} 
        L1+L2 & \multicolumn{2}{c|}{} & \multicolumn{2}{c}{For comparison}
    \end{tabular}
\end{table}

\subsection{Dataset}
To simulate human language acquisition, we trained models using child-directed speech (CDS) datasets from the CHILDES database~\cite{macwhinney-2001-childes} for both L1 and L2 training.
CDS datasets consist of recordings of naturalistic interactions between infants and their parents in home settings.
Specifically, we used the Providence corpus~\cite{borschinger-etal-2013-joint,lavechin-etal-2023-babyslm} for English, the MiiPro corpus~\cite{miyata-2012-guideline} for Japanese, and the Lyon corpus~\cite{demuth-tremblay-2008-prosodically} for French.
We applied three preprocessings to each dataset: removal of utterances by children, extraction of speech segments using voice activity detection (VAD)~\cite{lavechin-etal-2020-open}, and conversion to monaural format.
Additionally, we removed audio files without speaker annotation from the Lyon corpus.
All audio files were sampled at 16 kHz, and segments shorter than 8,000 samples were excluded from the datasets.
Table~\ref{tab:data-statistics} summarizes the statistics of the preprocessed datasets.
Each dataset was split into training and validation sets at a ratio of 99:1.
\begin{table}[tb]
  \centering
  \small
  \setlength{\tabcolsep}{2pt}
  \caption{Statistics of the dataset used. Parts. = Participants. Age = Age of Participants.}
  \label{tab:data-statistics}
  \begin{tabular}{clrlrrr}
    \toprule
    \textbf{Lang.} & \textbf{Dataset} & \textbf{\#Parts.} & \textbf{Age} & \textbf{\#Files} & \textbf{Duration} & \textbf{Avg. Duration} \\
    \midrule
    EN & Providence & 6 & 1-3 & 187,327 & 128.9h & 2.48s \\
    \midrule
    JA & MiiPro & 3 & 1-5 & 84,310 & 106.5h & 3.57s \\
    \midrule
    FR & Lyon & 5 & 1-3 & 83,449 & 81.6h & 3.52s \\
    \bottomrule
  \end{tabular}
\end{table}

\subsection{Model}
\begin{figure}[tb]
    \centerline{\includegraphics[width=9cm]{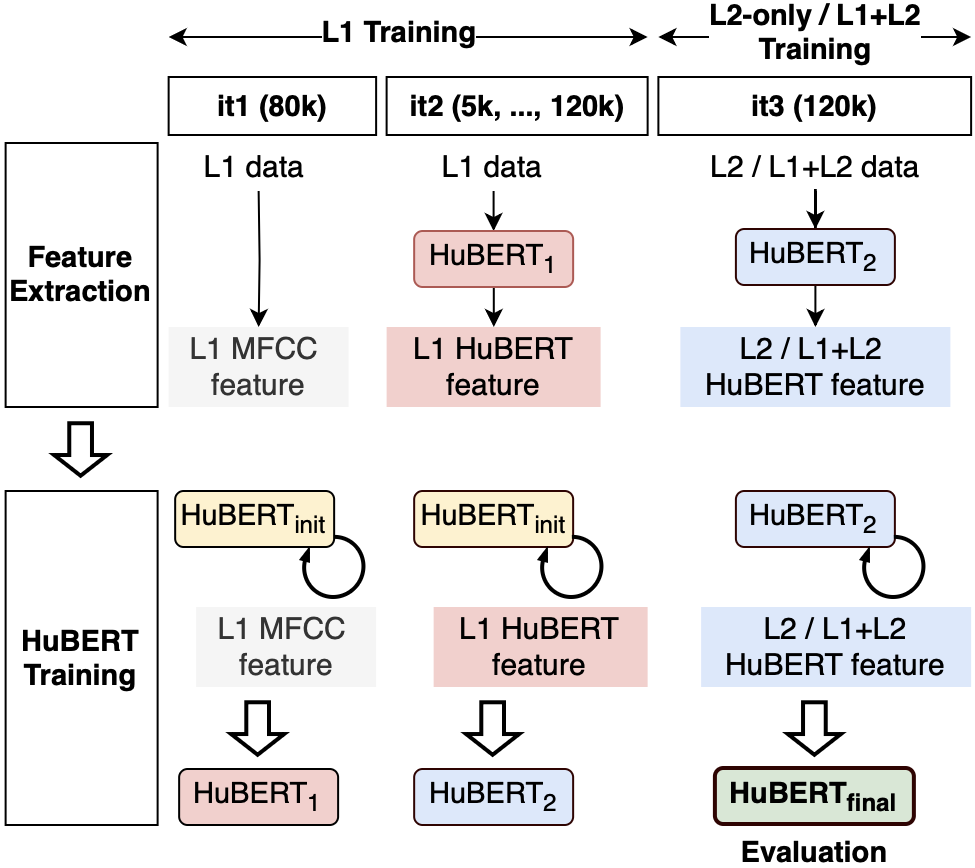}}
    \caption{Overview of the HuBERT training scheme. We train HuBERT for three iterations and test the CP effects by varying the number of L1 training steps in it2.}
    \label{fig:training}
\end{figure}

We adopted HuBERT~\cite{hsu-etal-2021-hubert} as the S3M model and conducted experiments using the Fairseq~\cite{ott-etal-2019-fairseq} implementation.
The models were trained in three iterations (it1, it2, it3) using different pseudo-labels, as illustrated in Fig.~\ref{fig:training}.
For each (L1, L2) pair, the total amount of data was matched across L1 and L2 by subsampling the larger dataset to the size of the smaller one (106.5h for English–Japanese pairs and 81.6h for English–French pairs).
All the models were trained with two different random seeds, and average scores are reported.

\subsubsection{L1 training (it1 and it2)}
We trained the model on L1 data for two iterations (it1, it2), following the original HuBERT training scheme.
It1 was trained for 80k steps and it2 for 120k steps (from scratch) to ensure near convergence in each iteration, as confirmed by validation loss.
We used pseudo-labels based on mel frequency cepstral coefficients (MFCC) in it1 and those derived from the features extracted from the it1 model in it2.
To train the models with various L2 exposure onsets and L1 exposure offsets, we saved the model checkpoints at 5k, 10k, 15k and every 15k steps thereafter during it2 (i.e., 30k, 45k, \ldots, 120k), and used these ten models as starting points for the subsequent L2 training.

\subsubsection{L2 training (it3)}
\begin{figure}[tb]
    \centerline{\includegraphics[width=9cm]{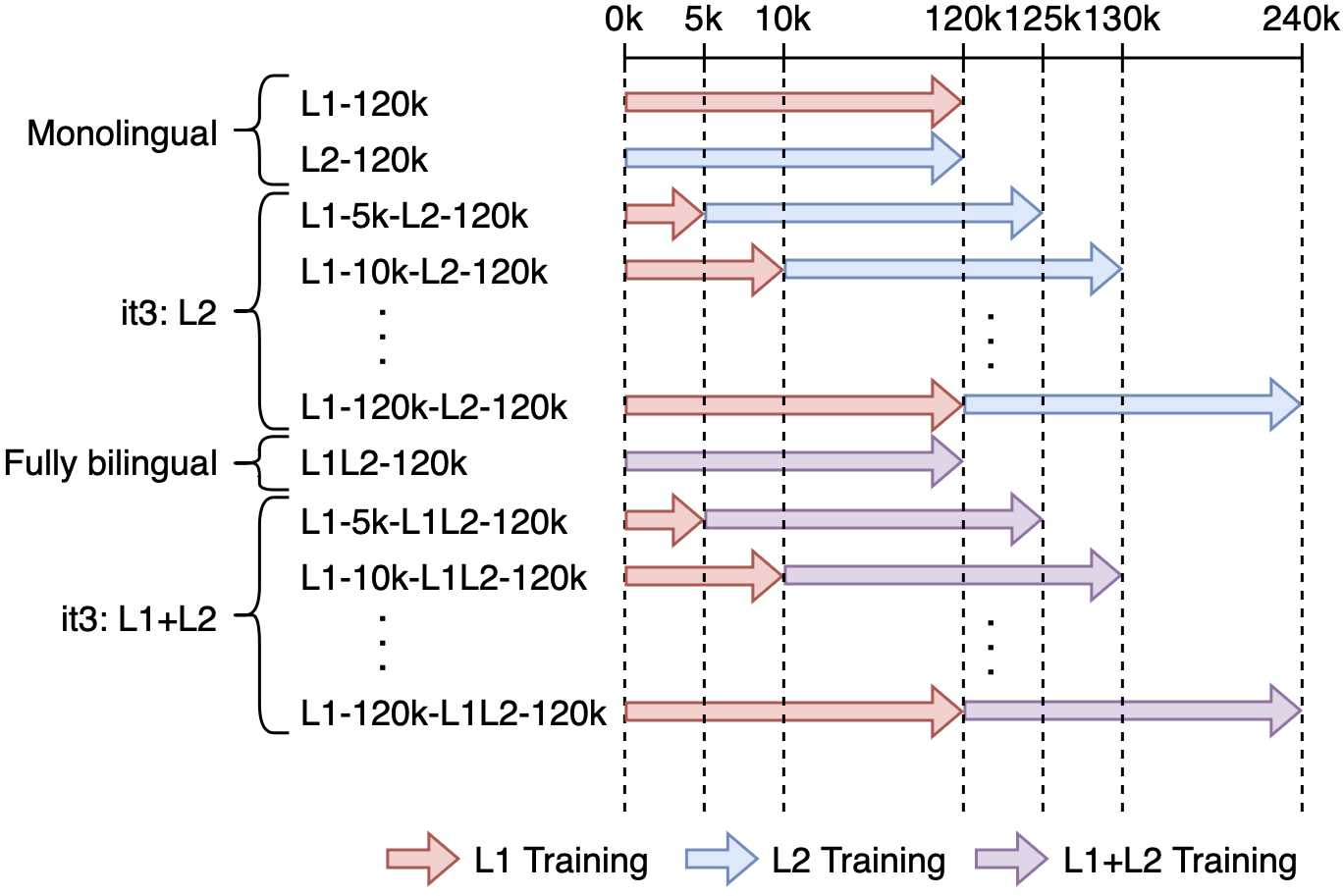}}
    \caption{Number of training steps for each model.}
    \label{fig:training-steps}
\end{figure}

From each of the ten L1-trained checkpoints saved in it2, we continued training on L2 for 120k steps (it3).
We considered two training settings here: training on only L2 as we call the L2-only setting, and training on a mixture of L1 and L2 data as we call the L1+L2 setting.
The purpose of the L1+L2 setting is to better reflect the human bilingual environment, since it is rare not to use L1 at all while learning L2.
As in it2, pseudo-labels in it3 were generated using representations from each L1-trained checkpoint.
The resulting ten models are referred to as L1-5k-L2-120k, L1-10k-L2-120k, \ldots, and L1-120k-L2-120k, respectively, according to the numbers of their L1 training steps.
The training durations of all the models are summarized in Fig.~\ref{fig:training-steps}.
Note that the number of L1 training steps can be regarded as the ones in it2 (5k, \ldots, 120k) without taking it1 training steps into account, since the models are trained from scratch in it2.

\subsubsection{Baselines}
As baselines, we trained monolingual models on L1 and L2 individually (L1-120k and L2-120k), and a fully bilingual model trained jointly on equal amounts of L1 and L2 (L1L2-120k) from scratch.
These models were trained for only two iterations (it1, it2); 80k steps in it1 and 120k steps in it2.

\subsection{Evaluation}
We evaluated the models' language acquisition performance with the English ABX phone discrimination test.
In the ABX test, three speech stimuli (A, B, and X) are presented and the task is to determine whether X is a stimulus of the same phoneme sequence as A or B.
Each of A, B, and X is a triphone, where A and B differ only in the central phoneme (e.g., (A, B, X) = (dig, dog, dig)), and X is spoken by a different speaker than A and B.
In the following, the stimulus (A or B) that matches X is referred to as \textit{target}, while the other stimulus is referred to as \textit{other}.
In this experiment, the models solve this task by comparing the distance between \textit{target} and X to that of \textit{other} and X.
Specifically, we extracted features for \textit{target}, \textit{other}, and X from HuBERT and compute the phone discrimination score $\Delta$ defined as follows~\cite{millet-dunbar-2022-self}:
\begin{align*}
    \Delta = DTW(M_{other},M_X) - DTW(M_{target},M_X),
\end{align*}
where DTW denotes the distance computed using dynamic time warping with cosine similarity, and $M_x$ represents the embedding computed by HuBERT for the input $x$.
This score serves as a measure of the model's ability to discriminate between the correct and incorrect options.
A positive $\Delta$ indicates that the model is correct, while a negative $\Delta$ indicates that it is incorrect.
We adopted accuracy as the evaluation metric.
We used the embeddings from the ninth Transformer layer of HuBERT, which is reported to encode phoneme-level information effectively~\cite{pasad-etal-2021-layer,sanabria-etal-2023-analyzing}.
As for the evaluation data, we used the ABX test in Zero Resource Speech Challenge 2017 (ZeroSpeech), which consists of 2,214 test triplets and is provided by the Perceptimatic database~\cite{millet-dunbar-2020-perceptimatic}.
\section{Results}
\subsection{CP effects for L2 acquisition}
\label{subsec:CP-L2acq}
\begin{figure*}[tb]
    \centerline{\includegraphics[width=16cm]{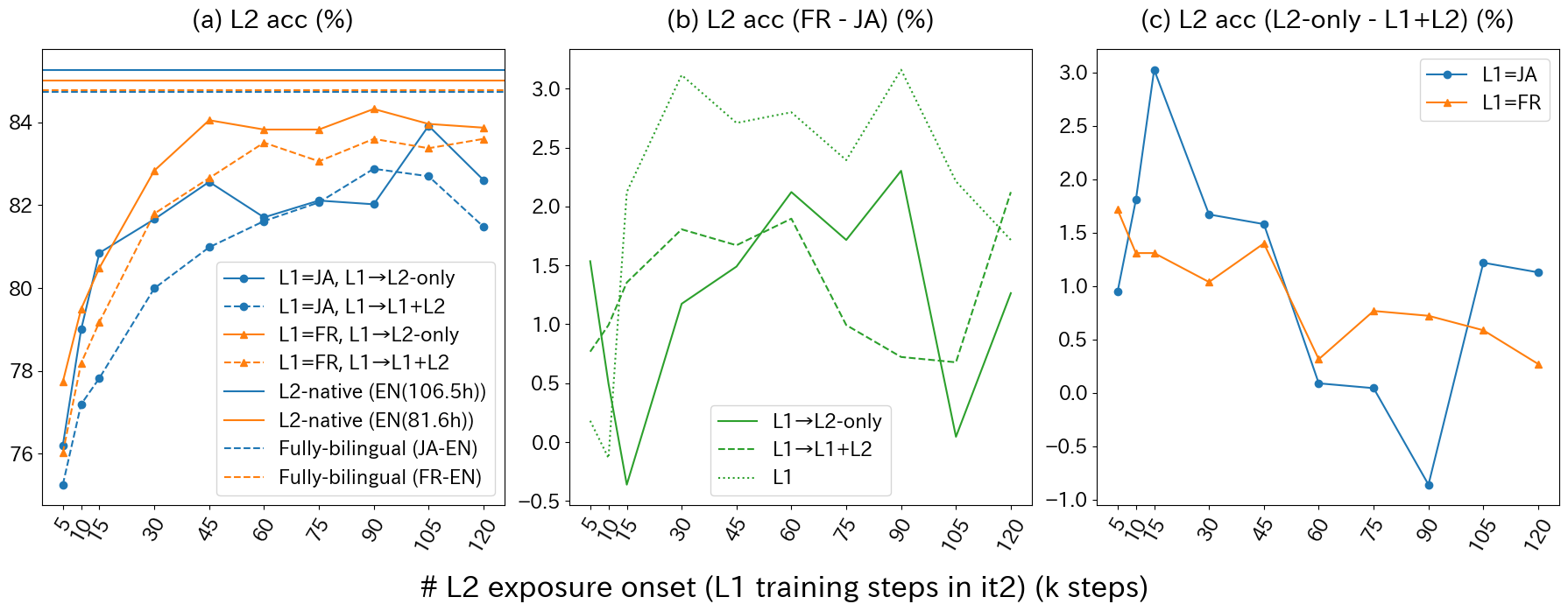}}
    \caption{\textbf{(a)}: L2 (EN) accuracies on the phone discrimination ABX test for models with various L1 training durations. We plot results for two L1s (Japanese and French), and for two L2 training settings with each L1: L2-only and L1+L2. The x-axis corresponds to each model (i.e., L1-5k-(L1)L2-120k, $\ldots$, and L1-120k-(L1)L2-120k). The horizontal lines show the L2 accuracies for the baseline models (L2-120k and L1L2-120k). \textbf{(b)}: L2 (EN) ABX accuracy advantage of training with French as L1 over Japanese, across two L2 training settings and the L1-120k setting. \textbf{(c)}: L2 (EN) ABX accuracy advantage of L2-only training in it3 over training jointly on L1 and L2, across two L1s.}
    \label{fig:CP-L2acq}
\end{figure*}

Fig.~\ref{fig:CP-L2acq} presents the L2 (English) ABX phone discrimination accuracy for models trained with English as an L2 (i.e., the left four settings in Table~\ref{tab:exp-setting}).
Fig.~\ref{fig:CP-L2acq} (a) shows the L2 accuracy for models with various L2 exposure onsets in each setting.
If the models exhibited CP effects for L2 acquisition similar to human learners, we would expect a monotonic decline in L2 performance with delayed L2 onset.
However, we observe the opposite trend: L2 performance generally improves or remains stable with more L1 training.
This pattern holds consistently across both L1 languages and both training settings in it3 (L2-only and L1+L2), suggesting that HuBERT does not exhibit CP effects for L2 acquisition under the current training regime and evaluation task.
Rather, this indicates that phone discrimination performance in one language improves with longer training regardless of the language used for training.
In addition, while humans exposed to an L2 in early childhood often achieve native-like proficiency, none of the models---including those with the earliest L2 exposure---reach the level of accuracy attained by the L2-native models.

\subsection{CP effects for L1 attrition}
\begin{figure*}[tb]
    \centerline{\includegraphics[width=16cm]{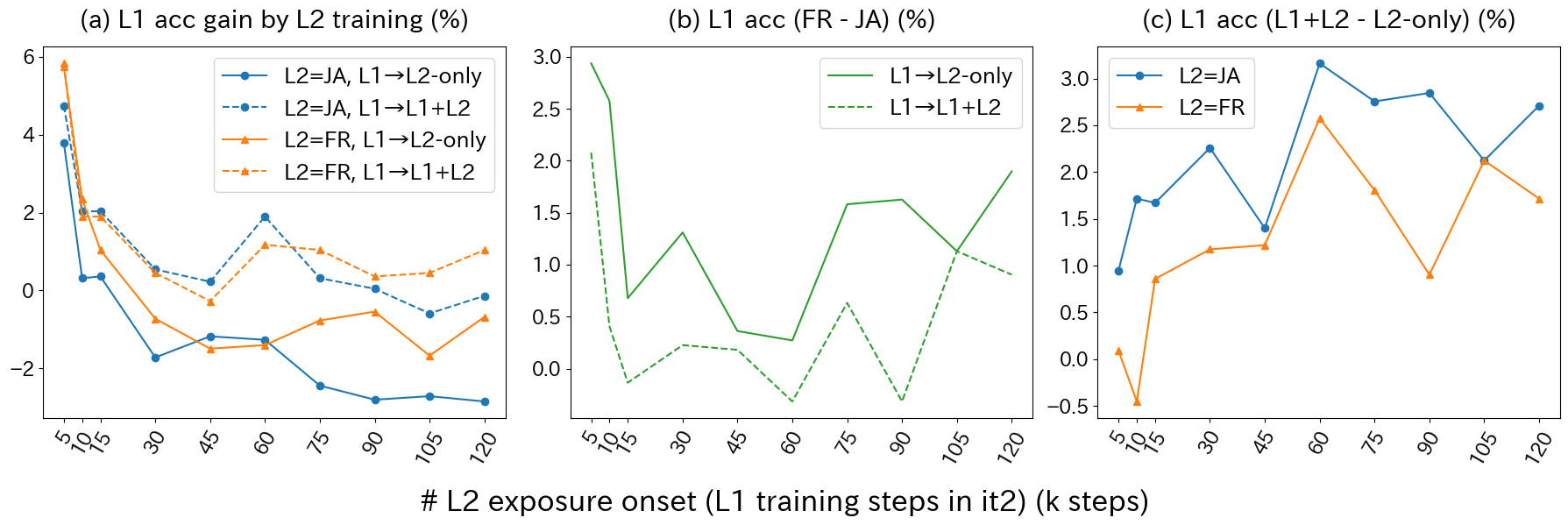}}
    \caption{\textbf{(a)}: L1 (EN) accuracy gains due to L2 training, computed by subtracting the L1 accuracy before L2 training from the one after L2 training. The x-axis can be regarded as the L1 exposure offset for L2-only models. \textbf{(b)}: L1 (EN) ABX accuracy advantage of training with French as L2 over Japanese. \textbf{(c)}: L1 (EN) ABX accuracy advantage of \textbf{training jointly on L1 and L2 over L2-only training} in it3.}
    \label{fig:CP-L1att}
\end{figure*}

Fig.~\ref{fig:CP-L1att} illustrates the L1 (English) ABX accuracy for all the models trained on English as their L1 (i.e., the right four settings in Table~\ref{tab:exp-setting}).
Fig.~\ref{fig:CP-L1att} (a) shows the L1 accuracy gain for each model caused by L2 training, computed by subtracting the pre-L2-training L1 accuracy from the final L1 accuracy, plotted against the L2 exposure onset.
If the models showed CP effects for L1 attrition, we would expect that L2-only models exhibit a monotonic increase in L1 performance gain with more L1 training steps (i.e., smaller L1 losses with a later L1 exposure offset).
However, our results show a different pattern.
L2-only models with a later L1 exposure offset tend to suffer greater degradation in L1 accuracy after L2 training.
In addition, those with an early L1 exposure offset show improved L1 accuracy with L2 training.
These results suggest that the L1 attrition patterns in HuBERT do not exhibit CP effects.
Rather than showing increased resistance to forgetting with prolonged L1 exposure, later L1 offset models appear more vulnerable to L1 interference from L2 training.

These results, along with those presented in Section~\ref{subsec:CP-L2acq}, are consistent with previous findings on textual language models~\cite{constantinescu-etal-2024-investigating}.
That study showed that text language models trained in a standard way without explicitly modeling neural plasticity do not exhibit the CP effects, either for L2 acquisition or L1 attrition.
Taken together, these findings suggest that CP phenomena are not inevitable outcomes of statistical learning alone, but instead may depend on innate neurobiological mechanisms.
It is also necessary to examine these effects on various tasks, such as the one that requires syntactic or semantic knowledge in addition to phonological knowledge.
Reverse-engineering the CP effects by modeling developmental characteristics of human brains will be a promising future direction for revealing the mechanism of this phenomenon.

\subsection{Analysis}
Here, we provide analyses of our results from other perspectives than the CP effects.

\subsubsection{Effects of L1 pretraining on L2 performance}
We first compare the L2 performance of the L2-native models (EN(106.5h)-120k, EN(81.6h)-120k) with the L1-\textit{n}k-L2-120k models, shown in Fig.~\ref{fig:CP-L2acq} (a).
These models are trained on L2 for 120k steps, and the only difference is that the latter models are pretrained on L1.
We observe that pretraining on a different language (L1) degrades L2 performance, indicating the occurrence of negative transfer.
This effect is more pronounced when the amount of L1 pretraining is limited, suggesting that insufficient L1 learning particularly hinders adaptation to L2.
Although increasing the amount of L1 pretraining improves L2 performance, it still does not match that of the L2 monolingual models.
A similar trend is observed when comparing L1L2-120k models with the L1-\textit{n}k-L1L2-120k models, trained jointly on L1 and L2 for 120k steps, with the same difference as above.
This pattern also holds regardless of whether the L1 is French or Japanese, which indicates that transfer learning in the ABX discrimination task may be inherently difficult even when the source and target languages are similar.
It also raises the possibility that training HuBERT for two iterations may be preferable over training for three iterations.
We incorporated this setting to create pseudo-labels for L2 data using L1-trained models, like the way monolingual humans perceive L2, but HuBERT is reported that the third iteration makes little difference to the performance when trained on only one language~\cite{hsu-etal-2021-hubert}.
Our results may reflect the inherent limitations of this training paradigm, which suggests that alternative training regimes (e.g., longer it2 training), different methods of label creation, or alternative models should be explored in future work.

\subsubsection{L2-only vs. L1+L2 training}
Here we analyze the effect of L2-only and L1+L2 training on both L2 and L1 performance.
Regarding L2 performance, as shown in Fig.~\ref{fig:CP-L2acq} (c), L2-only training generally leads to higher L2 accuracy than L1+L2 training, likely due to greater exposure to the target language.
In addition, a comparison between the L1L2-120k and L2(EN)-native baseline models reveals that bilingual models perform worse than monolingual models (see Fig.~\ref{fig:CP-L2acq} (a)).
In human studies, simultaneous bilinguals are often found to have lower phonological proficiency in their L2 compared to L2 monolinguals, which is commonly attributed to reduced L2 input~\cite{sebastian-galles-etal-2005-influence}.
Although these findings were based on Catalan and Spanish, similar trends have also been observed in grammatical proficiency in English across various L1 backgrounds~\cite{hartshorne-etal-2018-critical}.
Our findings align with these human studies, possibly due to the same reason.

Turning to L1 performance, L1+L2 training generally outperforms L2-only training, as shown in Fig.~\ref{fig:CP-L1att} (c).
This demonstrates that continued exposure to L1 helps preserve L1 performance, possibly due to the increased amount of exposure to L1.
This also supports findings from transfer learning literature, where jointly training on the previous and current tasks, known as the rehearsal methods in continual learning, is reported to mitigate catastrophic forgetting~\cite{rolnick-etal-2019-experience}.
Interestingly, for models that begin L2 training early, L1+L2 training sometimes results in lower L1 accuracy than L2-only training.
However, in these cases, L1 accuracy improves compared to the pre-L2 training stage (see Fig.~\ref{fig:CP-L1att} (a)), suggesting that no L1 forgetting occurs in these conditions.
In contrast, when L2 training begins late, L2-only models tend to suffer from L1 forgetting for both L1s, while L1+L2 training effectively prevents this degradation.
Thus, L1+L2 training is especially beneficial when L2 exposure is introduced after substantial L1 training.

\subsubsection{Effects of language similarity}
Fig.~\ref{fig:CP-L2acq} (b) shows the L2 accuracy advantage of using French over Japanese as the L1.
In most cases, models trained with French as their L1 outperform those trained with Japanese, despite the fact that the L2 training set (not the training steps) is smaller in size when trained on French.
This trend is consistent across varying L2 exposure onsets.
This advantage likely stems from the greater phonological similarity between French and English compared to Japanese and English, and this effect outweighs the effect of the dataset size.

As for the L1 accuracy advantage, although models trained with French as L2 perform better than those with Japanese in the L2-only setting, there is little or no advantage in the L1+L2 setting.

\section{Conclusion}
In this work, we investigated whether phonological CP effects for L2 acquisition and L1 attrition are observed in self-supervised speech models.
Specifically, we trained HuBERT varying the onset of L2 exposure and the offset of L1 exposure, and evaluated their phonological performance using a phone discrimination task.
We found that HuBERT does not exhibit either of the CP effects observed in humans: delayed L2 exposure onset improves L2 performance, delayed L1 exposure offset leads to L1 performance decline after L2 training, while early L1 exposure offset leads to improvement in L1 performance by L2 training.
These results indicate that the CP phenomenon is not an inevitable consequence of statistical learning alone, and imply the need for innate mechanisms to explain these effects.
A promising direction for future work involves examining tasks that require grammatical or semantic knowledge and reverse-engineering the CP effects, in order to better understand the learning mechanism of current models and the ones underlying the CP phenomenon.

\bibliographystyle{IEEEtran}
\bibliography{mybib}
\vspace{12pt}

\end{document}